\documentclass[sigconf]{acmart}
\AtBeginDocument{%
  }

\setcopyright{none}
\renewcommand\footnotetextcopyrightpermission[1]{}

\copyrightyear{2026}
\acmYear{2026}
\acmDOI{}     
\acmISBN{}    

\acmConference[LARS @ WWW '26]{LLM \& Agents for Recommendation Systems (LARS) Workshop}{2026}{TBD}
\acmBooktitle{Proceedings of the LLM \& Agents for Recommendation Systems (LARS) Workshop at The Web Conference 2026 (WWW '26 Companion)}

\usepackage{tabularx}
\usepackage{algorithm}
\usepackage{algpseudocode}
\usepackage{placeins}

\begin{document}


\title{PCN-Rec: Agentic Proof-Carrying Negotiation for Reliable Governance-Constrained Recommendation}

\author{Aradhya Dixit}
\affiliation{%
  \institution{Wake Technical Community College}
  \city{Raleigh}
  \state{North Carolina}
  \country{USA}
}
\email{adixit1@my.waketech.edu}

\author{Shreem Dixit}
\affiliation{%
  \institution{University of North Carolina at Charlotte}
  \city{Charlotte}
  \state{North Carolina}
  \country{USA}
}
\email{sdixit6@charlotte.edu}

\renewcommand{\shortauthors}{Dixit et al.}
\begin{abstract}
Modern LLM-based recommenders can generate compelling ranked lists, but they struggle to reliably satisfy governance constraints such as minimum long-tail exposure or diversity requirements. We present \textbf{PCN-Rec}, a \emph{proof-carrying negotiation} pipeline that separates natural-language reasoning from deterministic enforcement. A base recommender (MF/CF) produces a candidate window of size $W$, which is negotiated by two agents: a \emph{User Advocate} optimizing relevance and a \emph{Policy Agent} enforcing constraints. A mediator LLM synthesizes a Top-$N$ slate together with a structured \emph{certificate} (JSON) describing the claimed constraint satisfaction. A deterministic verifier recomputes all constraints from the slate and accepts only verifier-checked certificates; if verification fails, a deterministic constrained-greedy repair produces a compliant slate for re-verification, yielding an auditable trace.
On MovieLens-100K with governance constraints, PCN-Rec achieves a \textbf{98.55\%} pass rate on feasible users ($n=551$, $W=80$) versus a \textbf{one-shot} single-LLM baseline without verification/repair, while preserving utility with only a \textbf{0.021} absolute drop in NDCG@10 (\textbf{0.403} vs. \textbf{0.424}); differences are statistically significant ($p<0.05$).
\end{abstract}





\maketitle

\section{Introduction}

Recommender systems deployed in real platforms are governed by \emph{hard} constraints that go beyond pure relevance---for example, minimum long-tail exposure, genre diversity, or per-slate policy limits \cite{schedl2025torsspecialissue,jannach2023multiobjective,klimashevskaia2023popbias}. These constraints are often legally or contractually motivated, and must be \emph{auditable}: a platform must be able to explain why a slate was served and demonstrate that policy checks were actually satisfied. Recent LLM-based recommenders can produce persuasive rankings in natural language, but they are brittle when asked to reliably satisfy strict combinatorial requirements \cite{wu2024surveyllmrec,bao2023llmrecsyswww,li2024generativerec,geng2022p5,bao2023tallrec}. In practice, a single LLM may output a plausible justification while still violating constraints, and its reasoning is not directly executable or verifiable. Existing monolithic LLM approaches suffer from 'lost-in-the-middle' reasoning and an inability to maintain global state over combinatorial constraints \cite{zhu2023calibration}. By decomposing the recommendation task into an agentic competition, we allow for specialized reasoning paths: the User Advocate remains 'blind' to policy to maximize utility, while the Policy Agent provides a focused 'adversarial' check.

We propose \textbf{PCN-Rec}, which treats the LLM as a \emph{proposer} rather than an authority. The key idea is a \emph{proof-carrying} interaction: the LLM participates in negotiation, but every slate must pass a deterministic verifier implemented in code. PCN-Rec begins with a standard candidate generator (e.g., MF/CF) that produces a ranked list, from which we take a candidate window of size $W$. Within this window, two agents represent competing objectives: a \emph{User Advocate} pushes for relevance to the user’s preferences, while a \emph{Policy Agent} pushes for satisfaction of governance constraints (e.g., head/tail balance and genre diversity). A mediator LLM then proposes a Top-$N$ slate and emits a structured \emph{certificate} (JSON) describing which items satisfy which constraints. Crucially, a deterministic verifier recomputes all constraint functions from the proposed slate and either accepts the certificate or rejects it. When rejected, we apply a deterministic constrained-greedy repair to produce a compliant slate, re-verify, and log the trace for audit. Unlike monolithic LLM recommenders, our agentic approach decomposes the recommendation task into specialized roles, ensuring that the 'creative' proposer is decoupled from the 'strict' verifier.

\noindent\textbf{Contributions.} We make three contributions:
\begin{itemize}
  \item \textbf{Proof-carrying negotiation for recommendation:} an agentic fram where an LLM proposes slates and structured certificates, but correctness is enforced by deterministic verification \cite{wang2025verifiableformat,fan2024verifast}.
  \item \textbf{Feasibility analysis:} a window-size-based methodology that separates \emph{unsatisfiable} user histories (no feasible slate exists within $W$) from \emph{method failure} (a feasible slate exists, but the method fails).
  \item \textbf{Empirical evidence of strong governance with minimal utility loss:} on MovieLens-100K, PCN-Rec achieves near-perfect pass rate on feasible users with only a small NDCG@10 drop versus a single-LLM baseline \cite{harper2015movielens,jarvelin2002cg}.
\end{itemize}

\begin{figure*}[t]
  \centering
  \includegraphics[width=\textwidth]{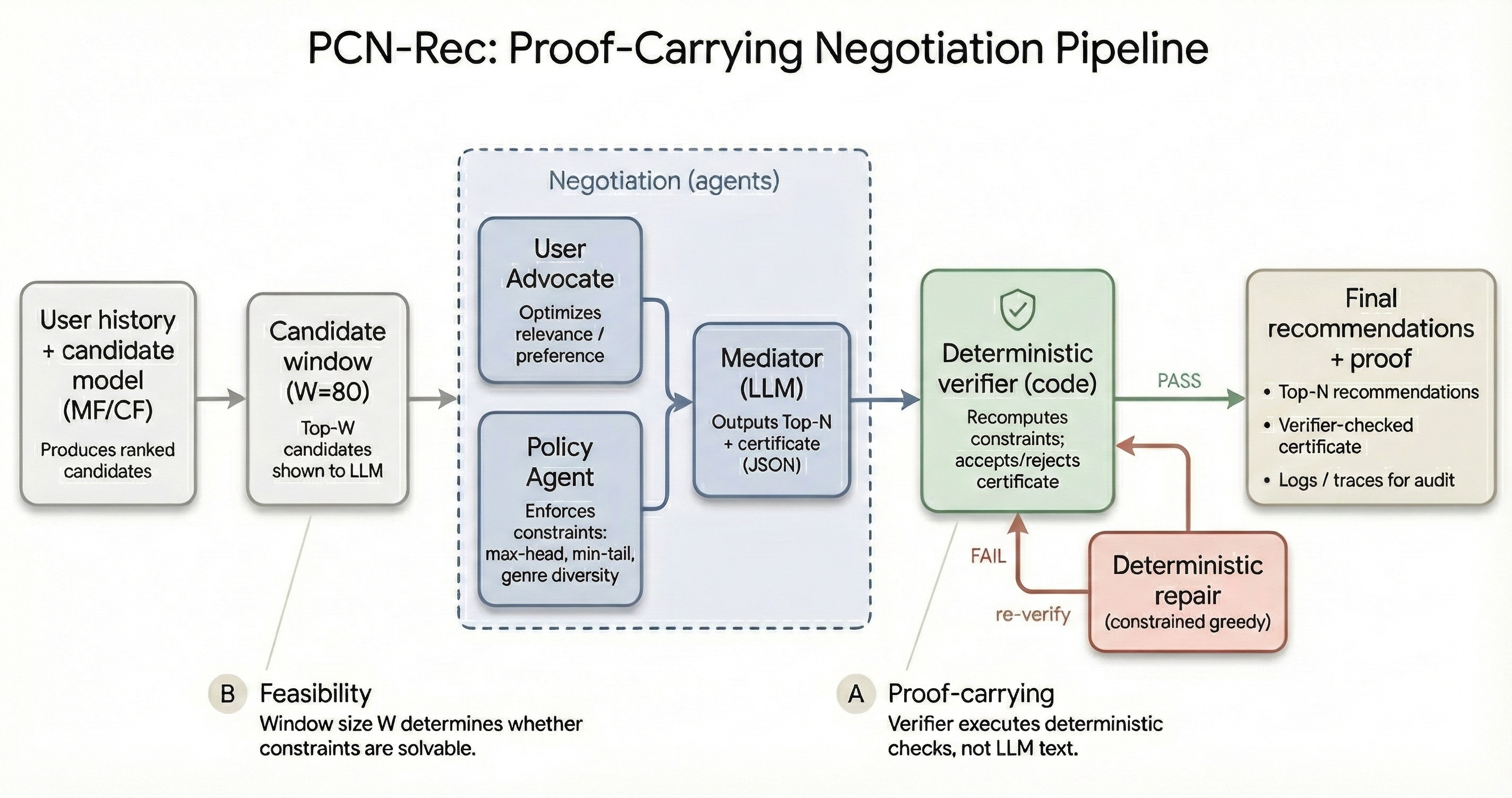}
  \caption{PCN-Rec proof-carrying negotiation pipeline. A base recommender produces a candidate window ($W$). Two agents negotiate (User Advocate for relevance; Policy Agent for governance), and a mediator LLM proposes a Top-$N$ slate plus a structured certificate. A deterministic verifier recomputes constraints from the slate and accepts only verifier-checked certificates; failures trigger deterministic constrained-greedy repair followed by re-verification, producing an auditable trace.}
  \Description{System diagram showing candidate generation, negotiation among agents with an LLM mediator emitting a structured certificate, deterministic verification, and a repair-and-reverify loop leading to final recommendations with an audit trail.}
  \label{fig:pipeline}
\end{figure*}

\section{Method: Proof-Carrying Negotiation for Recommendation}

Let $u$ denote a user, and let a base recommender (e.g., MF/CF) produce a ranked candidate list
$C(u)=\langle i_1,i_2,\dots\rangle$ with predicted relevance scores.
We restrict attention to a candidate window $C_W(u)$ containing the top-$W$ items, and aim to output a slate
$S \subseteq C_W(u)$ with $|S|=N$ that is both useful and compliant.

We consider governance constraints that must hold \emph{per slate} \cite{harper2015movielens,klimashevskaia2023popbias,bridge2024llmdiversity}. In our experiments, these include:
(i) a \textbf{head/tail exposure} constraint that enforces minimum long-tail representation and optionally caps head items,
and (ii) a \textbf{genre diversity} constraint that enforces a minimum number of distinct genres in the slate.
Each item $i$ is tagged with metadata (e.g., popularity bucket head/tail; genre labels) and constraints are computed
deterministically from these tags and the slate content.

\subsection{PCN-Rec Framework}
Figure~\ref{fig:pipeline} gives an overview. PCN-Rec separates natural-language reasoning from enforcement by treating
the LLM as a \emph{proposer} and placing all correctness guarantees in deterministic code \cite{wu2024surveyllmrec,bao2023llmrecsyswww,li2024generativerec,geng2022p5,bao2023tallrec}.

\paragraph{Candidate Window.}
The base recommender provides a ranked $C_W(u)$ that defines the feasible search space for all downstream steps.
This ``bounded'' view is important for auditability and for feasibility analysis: if no compliant slate exists within $C_W(u)$,
then failure is \emph{unsatisfiable constraints} rather than a method error.

\paragraph{Negotiation Agents.}
PCN-Rec uses two agents with complementary objectives\cite{huang2023recagent,zhang2025llmagentsurvey}:
\begin{itemize}
  \item \textbf{User Advocate:} argues for items that maximize relevance to the user’s history and preferences.
  \item \textbf{Policy Agent:} argues for items that satisfy governance constraints (e.g., min-tail, max-head, genre diversity).
\end{itemize}
Both agents operate only over candidates in $C_W(u)$.

\paragraph{Mediator (LLM) + Certificate.}
A mediator LLM synthesizes the agents’ arguments and proposes a Top-$N$ slate $S$, \emph{plus} a structured
\textbf{certificate} $c$ (JSON)\cite{lu2022neurologic,beurerkellner2022lmql,wang2025verifiableformat,michailidis2024constraintllms}.
The certificate contains (a) the selected item IDs, and (b) explicit fields needed to verify
governance constraints (e.g., which items are counted as tail; which genres are covered), enabling machine-checkable auditing.

\paragraph{Deterministic Verification.}
A verifier $V$ implemented in code recomputes all constraint functions directly from $S$ (and item metadata) and
accepts the output only if every constraint is satisfied \cite{fan2024verifast,lentz2025verifysoftware}. Critically, the verifier does \emph{not} trust LLM text ;
it trusts only deterministic checks. This yields a proof-carrying interface:
the system returns $(S,c)$ only when $V(S,c)=1$.

\paragraph{Deterministic Repair (Fail-Safe).}
If verification fails, PCN-Rec runs a deterministic constrained-greedy repair procedure that constructs a compliant slate
from $C_W(u)$ when possible (e.g., by first satisfying hard constraints, then maximizing relevance among remaining slots) \cite{ren2023slateaware,deffayet2023generativeslate,jannach2023multiobjective}.
The repaired slate is then re-verified and returned with an audit trace. This ensures robust compliance on feasible users
and cleanly separates \emph{constraint infeasibility} from \emph{LLM proposal failure}.

\subsection{Feasibility vs.\ Method Failure}
Because constraints are enforced per slate, some users may be infeasible under a given window size $W$.
We therefore report results on (i) all users and (ii) the subset of \textbf{feasible users} for which a compliant slate exists
within $C_W(u)$. This distinction is crucial for fair evaluation: infeasible cases should not be counted as algorithmic failures.
In Section~\ref{sec:feasibility} we quantify how feasibility changes with $W$ and use it to select the operating point $W{=}80$.


\section{Feasibility Analysis and Empirical Results}
\label{sec:feasibility}

We evaluate PCN-Rec on MovieLens-100K under per-slate governance constraints.
Our central question is: \emph{when are constraints satisfiable given a bounded candidate window}, and when failures are due to the method (e.g., LLM proposal errors) rather than true infeasibility.

\subsection{Problem Setup and Constraints}
Let $u$ denote a user. The base recommender returns a ranked candidate window
$C_W(u)=\langle i_1,\dots,i_W\rangle$ with predicted relevance scores $r_u(i)$.
A slate is an ordered list $S=(s_1,\dots,s_N)$ where $s_j \in C_W(u)$ and all items are distinct.
We write the set of feasible slates within the window as
\begin{equation}
\mathcal{S}_W(u) \;=\; \{ S \subseteq C_W(u) \;:\; |S|=N \}.
\end{equation}

Each item $i$ has metadata used by deterministic checks (e.g., popularity bucket and genre set).
Let $\mathbb{I}[\cdot]$ be the indicator function. We instantiate two common platform constraints:

\paragraph{Head/Tail exposure.}
Let $\mathrm{tail}(i)\in\{0,1\}$ indicate whether $i$ is long-tail (and $\mathrm{head}(i)=1-\mathrm{tail}(i)$).
We require at least a fraction $\tau$ of tail items and at most a fraction $\kappa$ of head items:
\begin{align}
\phi_{\text{tail}}(S) &:= \frac{1}{N}\sum_{i\in S}\mathrm{tail}(i) \;\ge\; \tau, \label{eq:tail}\\
\phi_{\text{head}}(S) &:= \frac{1}{N}\sum_{i\in S}\mathrm{head}(i) \;\le\; \kappa. \label{eq:head}
\end{align}

\paragraph{Genre diversity.}
Let $\mathcal{G}(i)$ be the set of genres for item $i$ and define the slate’s covered genres as
$\mathcal{G}(S)=\bigcup_{i\in S}\mathcal{G}(i)$. We require
\begin{equation}
\phi_{\text{div}}(S) \;:=\; |\mathcal{G}(S)| \;\ge\; G_{\min}. \label{eq:div}
\end{equation}

\subsection{Feasibility vs.\ Method Failure}
A key distinction in governance-constrained recommendation is whether constraints are \emph{unsatisfiable} within the candidate window.
We define window-feasibility for a user as
\begin{equation}
\mathrm{Feasible}_W(u) \;=\; \mathbb{I}\!\left[\exists S\in\mathcal{S}_W(u)\;:\;
\phi_{\text{tail}}(S)\wedge \phi_{\text{head}}(S)\wedge \phi_{\text{div}}(S)\right]. \label{eq:feasible}
\end{equation}
This separates \textbf{unsatisfiable} cases ($\mathrm{Feasible}_W(u)=0$) from \textbf{method failures} where a feasible slate exists but the proposer fails to find it.

Operationally, the strongest notion of “best possible” compliant slate is the constrained optimization
\begin{equation}
S_u^\star \;=\; \arg\max_{S\in\mathcal{S}_W(u)} \sum_{i\in S} r_u(i)
\quad \text{s.t.}\quad \eqref{eq:tail},\eqref{eq:head},\eqref{eq:div}. \label{eq:constrained-opt}
\end{equation}
PCN-Rec approximates \eqref{eq:constrained-opt} via (i) LLM-mediated negotiation and (ii) a deterministic repair fallback.

\begin{figure}[t]
  \centering
  \includegraphics[width=\linewidth]{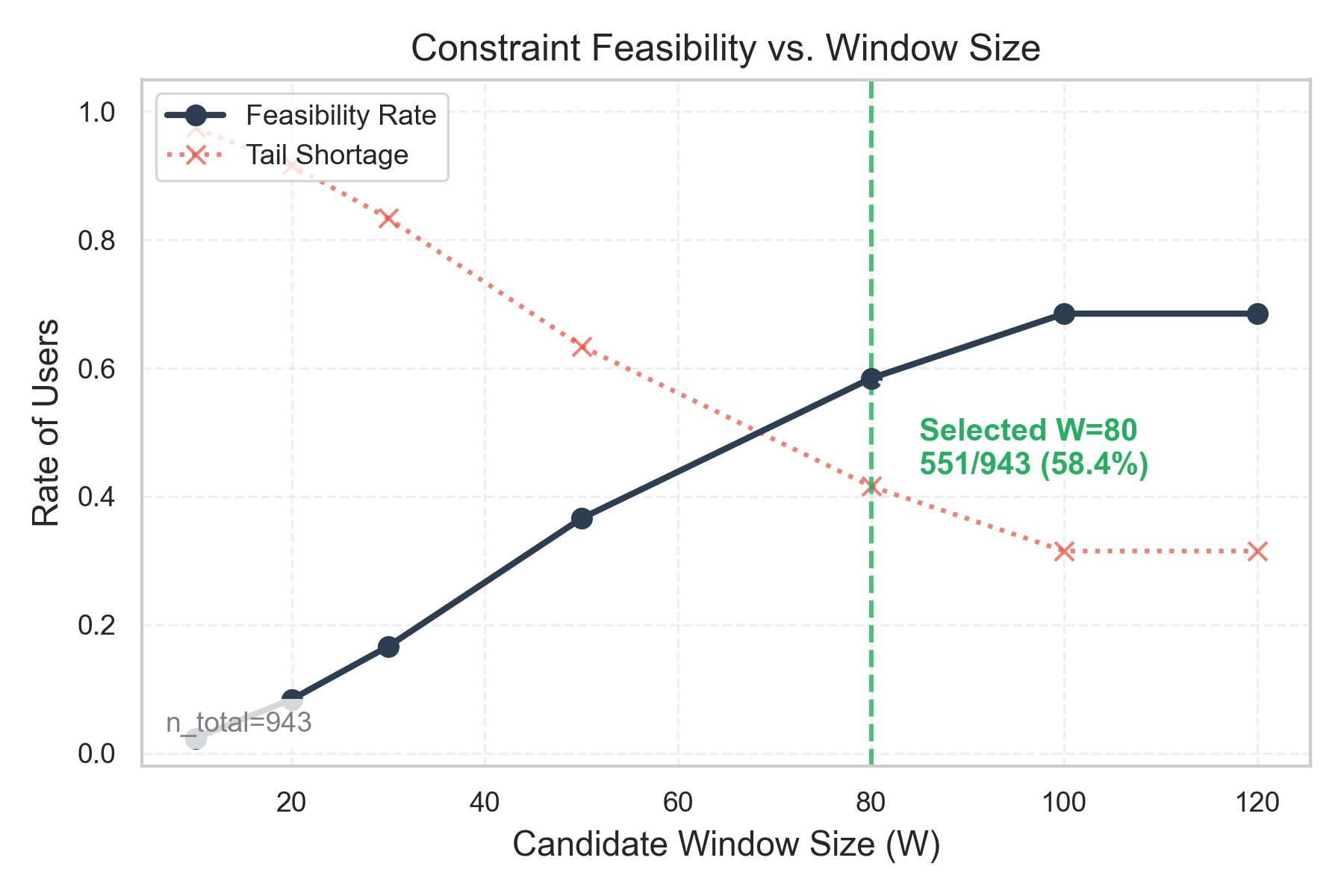}
  \caption{Constraint feasibility vs.\ candidate window size $W$. The feasibility rate (fraction of users for whom at least one compliant slate exists within $C_W(u)$) increases with $W$, while tail-shortage decreases. We select $W{=}80$ as an operating point.}
  \Description{Line plot showing feasibility rate increasing with window size and tail-shortage decreasing; a vertical marker at W=80 indicates the selected operating point.}
  \label{fig:feasibility}
\end{figure}

\subsection{Proof-Carrying Verification Metric}
Let the mediator produce a proposed slate $S$ and a structured certificate $c$ (JSON).
The verifier recomputes constraint functions from code and accepts only if both (a) the certificate is well-formed and consistent with $S$,
and (b) all governance constraints pass:
\begin{equation}
V(S,c) \;=\; \mathbb{I}\!\left[\mathrm{parse}(c)=S \;\wedge\; \bigwedge_{m\in\{\text{tail,head,div}\}} \phi_m(S)\right]. \label{eq:verifier}
\end{equation}

We report the governance pass rate as
\begin{equation}
\mathrm{PassRate} \;=\; \frac{1}{|\mathcal{U}|}\sum_{u\in\mathcal{U}} V(S_u,c_u),
\qquad
\mathrm{PassRate}_{\text{feas}} \;=\; \frac{1}{|\mathcal{U}_W^{+}|}\sum_{u\in\mathcal{U}_W^{+}} V(S_u,c_u),
\label{eq:passrate}
\end{equation}
where $\mathcal{U}_W^{+}=\{u\in\mathcal{U}:\mathrm{Feasible}_W(u)=1\}$.

\subsection{Utility Metric (NDCG@10)}
We measure ranking utility with NDCG@10 \cite{jarvelin2002cg}. For a ranked list $(s_1,\dots,s_K)$, define
\begin{align}
\mathrm{DCG@}K(u) &= \sum_{j=1}^{K}\frac{2^{\mathrm{rel}_u(s_j)}-1}{\log_2(j+1)}, \\
\mathrm{NDCG@}K(u) &= \frac{\mathrm{DCG@}K(u)}{\mathrm{IDCG@}K(u)},
\end{align}
where $\mathrm{IDCG@}K(u)$ is the DCG of the ideal ranking for user $u$.

\subsection{Main Findings}
Figure~\ref{fig:feasibility} shows how increasing $W$ increases feasibility (more users admit at least one compliant slate)
while reducing tail shortage. We select $W{=}80$ as an operating point that yields a large feasible subset ($551/943$ users)
without expanding the candidate set excessively.

On the feasible subset $\mathcal{U}_{80}^{+}$, PCN-Rec achieves near-perfect governance compliance (verifier-checked slates)
while incurring only a small drop in utility compared to a single-LLM baseline (Table~\ref{tab:gov-utility}).
This illustrates the core tradeoff: \emph{enforce governance deterministically, and treat the LLM as a proposer whose output must carry a checkable proof.}


\subsection{Governance Compliance vs.\ Utility Tradeoff}
\label{sec:gov-utility}

We compare PCN-Rec to a \emph{Single LLM} baseline that directly synthesizes a Top-$N$ slate without deterministic verification.
On the feasible subset $\mathcal{U}^{+}_{80}$, PCN-Rec achieves near-perfect governance compliance: verifier-checked slates pass at
$0.985$ (vs.\ $0.000$ for Single LLM), demonstrating that deterministic verification eliminates brittle constraint violations.
This compliance comes with a small utility reduction: NDCG@10 decreases from $0.424$ to $0.403$ ($\Delta=-0.022$),
consistent with the expected cost of enforcing hard governance constraints.
We also report a deterministic constrained-greedy bound as a reference point for the best achievable compliant utility under the same constraint set. Single LLM is evaluated as a one-shot proposer over the same candidate window $C_W(u)$, without verification or repair; PCN-Rec adds a verifier-checked certificate interface and a deterministic fail-safe repair path.

\begin{table}[t]
  \caption{Governance compliance and utility on feasible users ($n{=}551$, $W{=}80$).}
  \label{tab:gov-utility}
  \centering
  \small
  \setlength{\tabcolsep}{4pt}
  \renewcommand{\arraystretch}{1.1}

  \begin{tabularx}{\columnwidth}{@{}Xcc@{}}
    \toprule
    \textbf{Method} & \textbf{PassRate (feasible)} & \textbf{NDCG@10} \\
    \midrule
    Single LLM (no verifier) & 0.000 & 0.424 \\
    PCN-Rec (verifier-checked) & 0.985 & 0.403 \\
    Deterministic greedy bound (reference) & --- & 0.426 \\
    \bottomrule
  \end{tabularx}

  \vspace{2pt}
  {\footnotesize Utility drop: $\Delta\,\mathrm{NDCG@10} = -0.022$ (PCN-Rec vs.\ Single LLM).}
\end{table}


\paragraph{Statistical testing.}
To assess whether the utility difference is robust, we use a paired test over users in $\mathcal{U}^{+}_{80}$ (e.g., paired bootstrap over NDCG@10 per user).
Unless stated otherwise, we report significance at $p<0.05$.

\FloatBarrier

\section{Discussion and Related Work}
\label{sec:related}

\paragraph{Why proof-carrying negotiation?}
A practical pain point in governance-constrained recommendation is that platform policies are \emph{hard} (must always hold) and often \emph{auditable} (must be explainable after the fact). PCN-Rec operationalizes this by making the LLM a \emph{proposer} and placing correctness in deterministic code: every served slate is accompanied by a machine-checkable certificate and an explicit pass/fail decision from the verifier. This shifts compliance from “best-effort” to “guaranteed on feasible users,” and provides a direct artifact for auditing and debugging.

\begin{algorithm}[t]
\caption{PCN-Rec inference with proof-carrying verification and deterministic repair.}
\label{alg:pcnrec}
\begin{algorithmic}[1]
\Require User $u$, candidate window $C_W(u)$, slate size $N$, policy constraints $\{\phi_k\}_{k=1}^K$
\Ensure Compliant slate $S$ with auditable trace (certificate + verifier logs)
\State $C \gets \textsc{BaseRank}(u)$ \Comment{e.g., MF/CF scores}
\State $C_W \gets \textsc{TopW}(C, W)$
\State $(S, c) \gets \textsc{Negotiate}(u, C_W, N)$
\Comment{LLM mediator + agents output slate and structured certificate}
\If{$\textsc{Verify}(S, c, \{\phi_k\})$}
    \State \Return $(S, c, \textsc{Log}(\texttt{PASS}))$
\Else
    \State $S' \gets \textsc{RepairGreedy}(u, C_W, N, \{\phi_k\})$
    \State $c' \gets \textsc{MakeCert}(S', \{\phi_k\})$ \Comment{deterministic}
    \If{$\textsc{Verify}(S', c', \{\phi_k\})$}
        \State \Return $(S', c', \textsc{Log}(\texttt{FAIL}\rightarrow\texttt{REPAIR}\rightarrow\texttt{PASS}))$
    \Else
        \State \Return $(\bot, \bot, \textsc{Log}(\texttt{INFEASIBLE within }C_W(u)))$
    \EndIf
\EndIf
\end{algorithmic}
\end{algorithm}

\paragraph{Relation to constrained recommendation.}
Prior approaches to constrained recommendation and constrained ranking often rely on optimization-based re-ranking, heuristics, or post-processing to satisfy constraints while preserving relevance \cite{ren2023slateaware,deffayet2023generativeslate,jannach2023multiobjective,klimashevskaia2023popbias,ren2022mitigatepopbias,bridge2024llmdiversity}. PCN-Rec is compatible with this line of work (our repair is a deterministic constrained-greedy procedure), but differs in its \emph{interface contract}: the system returns an explicit proof object (certificate) that must validate under a deterministic verifier, making compliance transparent and reproducible \cite{ren2023slateaware,deffayet2023generativeslate,jannach2023multiobjective}.

\paragraph{Relation to LLM-based and agentic recommenders.}
LLM-based recommenders can produce natural-language rationales and adaptively perform multi-objective alignment preferences, but they are brittle under strict combinatorial constraints. PCN-Rec uses agents and an LLM mediator for flexibility and personalization, while making constraint satisfaction independent of the LLM’s textual reasoning. This separation keeps the interface benefits of LLMs while preventing silent constraint violations \cite{wu2024surveyllmrec,bao2023llmrecsyswww,li2024generativerec,huang2023recagent,zhang2025llmagentsurvey}.

\paragraph{Limitations.}
First, feasibility depends on the candidate window: if constraints are unsatisfiable within $C_W(u)$, no method can produce a compliant slate without expanding the window or relaxing constraints. Second, the verifier only checks \emph{codified} constraints (e.g., tail exposure, diversity); subjective or evolving policies still require careful formalization. Third, certificate schemas and metadata quality matter: incorrect item metadata can cause either false rejections or false acceptances unless the metadata framework is trusted and monitored.

\paragraph{Future work.}
Natural extensions include (i) richer constraint families (e.g., fairness constraints across groups, multi-slate exposure budgets), (ii) adaptive selection of $W$ per user to reduce infeasibility while controlling compute, and (iii) tighter coupling between repair and utility optimization (e.g., approximation guarantees or learned repair policies that remain verifier-safe).

\section{Conclusion}
We presented \textbf{PCN-Rec}, a proof-carrying negotiation framework for governance-constrained recommendation.
The core design choice is to treat the LLM as a \emph{proposer} that can negotiate relevance and policy tradeoffs,
while enforcing correctness via a deterministic verifier that recomputes all constraints in code.
This yields an auditable interface: every served slate is accompanied by a structured certificate and an explicit
verifier decision.

Empirically, we show that governance failures should be separated into (i) \emph{true infeasibility} within a bounded
candidate window and (ii) \emph{method failure} when a feasible slate exists but the proposer fails.
On MovieLens-100K under per-slate constraints, PCN-Rec achieves near-perfect governance compliance on feasible users \cite{harper2015movielens}
with only a small NDCG@10 reduction, illustrating a practical recipe for deploying LLM-driven recommenders without
silent policy violations.

\section{Limitations, Ethics, and Reproducibility}
\label{sec:limitations}

\subsection{Limitations}
PCN-Rec guarantees compliance only \emph{conditional on feasibility} within the candidate window $C_W(u)$.
If constraints are unsatisfiable for a user at the chosen $W$, no negotiation strategy can produce a compliant slate
without expanding the window, relaxing constraints, or changing the candidate generator.
Second, the verifier is only as good as the \emph{formalized} constraints and the \emph{metadata quality}
(e.g., tail/head labels and genre tags); errors in metadata can lead to incorrect accept/reject decisions.
Third, while deterministic repair is robust, it can introduce a utility gap versus globally optimal constrained ranking;
closing this gap while remaining verifier-safe is an important direction.

\subsection{Ethics and Governance Considerations}
Our goal is to improve platform accountability: policies that must hold (e.g., exposure constraints) are enforced by code
and logged for audit. However, governance constraints can be misused or encode value judgments (e.g., what counts as
``diversity'' or ``long-tail''), so stakeholders should document policy intent and validate metrics against real outcomes.
Audit logs and certificates can also reveal sensitive preference signals if stored improperly; deployments should minimize
logged personal data, apply access controls, and consider aggregation/anonymization.
Finally, LLM components may produce persuasive but misleading rationales; PCN-Rec mitigates \emph{silent} policy violations,
but it does not automatically guarantee that explanations are faithful unless explanation claims are also verified.

\subsection{Reproducibility Checklist}
To support reproducibility and auditing, we recommend the following artifacts and reporting:
\begin{itemize}
  \item \textbf{Data and splits:} MovieLens-100K version, preprocessing steps, train/test split protocol, and evaluation users.
  \item \textbf{Candidate generation:} base model (e.g., MF/CF), hyperparameters, and the exact definition of $C_W(u)$.
  \item \textbf{Constraint definitions:} explicit formulas and threshold values ($\tau$, $\kappa$, $G_{\min}$), plus how head/tail and genres are assigned.
  \item \textbf{Verifier + certificate schema:} JSON schema fields, parsing rules, and the deterministic verification code path.
  \item \textbf{LLM setup:} model name/version, decoding settings, system prompts for agents/mediator, and rate limits/timeouts.
  \item \textbf{Deterministic repair:} algorithm details (ordering, tie-breaking), and any randomness controlled via fixed seeds.
  \item \textbf{Metrics and tests:} NDCG@10 computation details and the paired significance test procedure.
\end{itemize}

Overall, PCN-Rec provides a lightweight, deployable pattern for combining LLM flexibility with deterministic governance guarantees.


\bibliographystyle{ACM-Reference-Format}
\bibliography{sample-base}










\end{document}